\documentclass[conference]{IEEEtran}

\ifCLASSINFOpdf
\else
   \usepackage[dvips]{graphicx}
\fi
\usepackage{url}

\usepackage{graphicx}
\usepackage{amsmath}
\usepackage{bm}
\usepackage{algorithm}
\usepackage{algorithmicx}
\usepackage{algpseudocode}
\usepackage{stfloats}
\usepackage{cuted}
\usepackage{capt-of}
\usepackage{hyperref}

\begin{document}
\pagestyle{plain}
\thispagestyle{plain}

\title{Fidelity-Constrained Anchoring for Black-Box Denoisers}
\author{\IEEEauthorblockN{Masaki Satoh}
\IEEEauthorblockA{
\textit{Morpho, Inc.}\\
Tokyo, Japan \\
m-satoh@morphoinc.com}
}

\maketitle

\IEEEpeerreviewmaketitle

\begin{abstract}
We propose a fidelity-constrained framework that anchors the output of a black-box denoiser to its input without retraining and with little additional computation. The method linearly blends the denoised image with the input and selects the maximum blending factor that satisfies a prescribed local fidelity constraint using Peak Signal-to-Noise Ratio (PSNR) or Structural Similarity Index (SSIM). For PSNR control, a closed-form solution is obtained under a local constant-blending assumption. For SSIM control, we derive a tractable formulation based on inverse SSIM under the same assumption and solve it efficiently using iterative root finding. Experiments on DIV2K images with synthetic Gaussian noise and outputs from Real-ESRGAN and a non-local means denoiser show that the proposed anchoring strategy provides effective fidelity control while balancing denoising performance and statistical naturalness, as measured by the excess kurtosis of residual noise. In particular, SSIM-based anchoring yields more consistent behavior across noise levels than PSNR-based anchoring.
\end{abstract}

\section{Introduction}

Denoising is a fundamental task in image processing \cite{morikawa2021}. Its objective is to improve perceptual quality by removing unwanted noise while preserving structural details. Although deep learning-based denoisers have recently achieved state-of-the-art performance, they often introduce unnatural artifacts. High-fidelity restoration is essential in applications such as surveillance. In this context, fidelity is defined as the similarity between a denoised output and its input image. Representative fidelity metrics include Peak Signal-to-Noise Ratio (PSNR) and the Structural Similarity Index (SSIM) \cite{wang2004image}. These metrics can be incorporated into training losses to improve fidelity. However, such training-time constraints do not guarantee that every output will satisfy a prescribed fidelity condition. Incorporating a denoiser into a feedback loop to enforce fidelity constraints is also possible, but this approach is impractical because directly controlling the output of a complex denoiser is difficult and may require repeated denoiser evaluations.

In this paper, we propose a framework that anchors the output of a black-box denoiser to the input image using fidelity metrics. Avoiding direct manipulation of the denoiser, we linearly blend the denoiser output $B$ with the input image $I$ to produce the output $O$:
\begin{align}
    O = I + \alpha \odot ( B - I ),
\end{align}
where $\alpha$ is a blending matrix and $\odot$ denotes element-wise multiplication. As $\alpha$ increases, $O$ approaches $B$ and fidelity to the input decreases. The blending factor $\alpha$ is computed to satisfy a fidelity condition between $I$ and $O$:
\begin{align}
    \hat{\alpha} = \max\{\alpha\,|\,f(I,O) \geq T \},
\end{align}
where $f(I,O)$ is the fidelity metric and $T$ is a prescribed threshold. In other words, we seek an operating point that balances fidelity and denoising performance. Because the framework is based on blending, image-quality tuning can be performed without retraining the denoiser, and the proposed method is applicable to a broad range of denoisers.

As fidelity metrics, we consider PSNR and SSIM. Because the Mean Squared Error (MSE) constraint corresponding to PSNR is quadratic in $\alpha$, we can easily obtain a closed-form solution. In contrast, SSIM is highly nonlinear in $\alpha$, but we show that its inverse has a tractable form and that the operating point can be found efficiently with a small number of iterations.

\section{Fidelity Metrics}
\label{sec:fidelity_metrics}

In this section, we consider two widely used metrics, PSNR and SSIM, and examine how to control them by adjusting the blending factor $\alpha$.

\subsection{PSNR}

The PSNR between two images $I$ and $O$ is defined as follows:
\begin{align}
    \mathrm{PSNR}(I,O) \equiv 10 \log_{10} \frac{L^2}{\mathrm{MSE}(I,O)},
\end{align}
where $L$ is the maximum possible pixel value, and $\mathrm{MSE}(I,O)$ is the mean squared error between $I$ and $O$. For 8-bit images, $L=255$. Therefore, PSNR is monotonically related to MSE and can be controlled through MSE. For simplicity, we consider MSE instead of PSNR, and the problem can be formulated as
\begin{align}
    \hat{\alpha} = \max\{\alpha\,|\, \mathrm{MSE}(I,O) \le T_\mathrm{MSE} \},
\end{align}
where $T_\mathrm{MSE}$ is a threshold on MSE.

MSE is usually computed over the entire image. However, this is too coarse for our purpose, so we compute MSE locally. Specifically, we compute MSE over local subimages of $I$ and $O$ centered at each pixel and impose the constraint on each subimage. We then assign the blending factor computed for each subimage to its corresponding center pixel. Let $i^k$, $o^k$, and $b^k$ be the $k$-th subimages of $I$, $O$, and $B$, respectively. We assume that the blending factor $\alpha$ takes a constant value $\alpha^k$ within each subimage. This working assumption may cause the fidelity condition to be violated in some cases, and we evaluate its validity in the experimental results. Then, the MSE between subimages $i^k$ and $o^k$ is
\begin{align}
    \mathrm{MSE}^k(\alpha^k) = (\alpha^k)^2 \mathrm{MSE}(i^k,b^k),
\end{align}
and the desired value of $\alpha^k$ is
\begin{align}
    \hat{\alpha}^k = \min\left\{1, \sqrt{\frac{T_\mathrm{MSE}}{\mathrm{MSE}(i^k,b^k)}}\right\}.
    \label{eq:alpha_mse}
\end{align}
Thus, MSE, and therefore PSNR, can be controlled by alpha blending under the constant-$\alpha$ assumption.

The MSE between $i^k$ and $b^k$ is expected to be sensitive to the noise level. When the input noise is sufficiently large, the MSE is expected to be large, making the blending factor $\alpha$ small, so the output $O$ almost coincides with $I$. In contrast, when the input noise is sufficiently small, $O$ almost coincides with $B$. Thus, we should tune the anchoring condition according to the input noise level. However, this is impossible because the input noise level is unknown in practice. In the following subsection, we show that SSIM is expected to be more robust to noise-level variations.

\subsection{SSIM}

In contrast to PSNR, SSIM is more difficult to control because of its nonlinear form. The SSIM between $I$ and $O$ is defined as follows:
\begin{align}
    \mathrm{SSIM}(I,O) \equiv
    \frac{(2\mu_{I} \mu_{O} + C_1)(2\sigma_{I O} + C_2)}
    {(\mu_{I}^2 + \mu_{O}^2 + C_1)(\sigma_{I}^2 + \sigma_{O}^2 + C_2)},
\end{align}
where $\mu_{I}$ and $\mu_{O}$ are the mean intensities of $I$ and $O$, $\sigma_{I}^2$ and $\sigma_{O}^2$ are the variances, $\sigma_{I O}$ is the covariance, and $C_1$ and $C_2$ are stabilizing constants. Typical values for 8-bit grayscale images are $C_1 = (0.01\times255)^2$ and $C_2 = (0.03\times255)^2$. When $I$ and $O$ are identical, $\mathrm{SSIM}(I,O)=1$; as their difference increases, $\mathrm{SSIM}(I,O)$ decreases toward $0$. If two images are inversely correlated, SSIM can take negative values.

In practice, the means and (co)variances are computed with small Gaussian windows to obtain local SSIM maps, which are then averaged to produce a single global SSIM value. We impose the fidelity condition on each local SSIM value. For simplicity, we approximate Gaussian-window convolution by considering subimages centered at each pixel, as in the PSNR case. Assuming that the blending factor $\alpha$ takes a constant value $\alpha^k$ within each subimage, we define the inverse local SSIM as follows:
%
\begin{align}
    \psi^k(\alpha) \equiv \frac{1}{\mathrm{SSIM}^k(\alpha)} \equiv \psi^k_\mu(\alpha) \psi^k_\sigma(\alpha),
\end{align}
where $\psi^k_\mu(\alpha)$ and $\psi^k_\sigma(\alpha)$ are the $\mu$- and $\sigma$-dependent components of inverse SSIM, respectively, and are given by
\begin{align}
    \psi^k_\mu(\alpha) &= 1
    + \frac{\alpha^2 (\mu_{b} - \mu_{i})^2}{2\mu_{i}^2 + 2 \alpha \mu_{i} (\mu_{b} - \mu_{i}) + C_1} \\
    \psi^k_\sigma(\alpha) &= 1
    + \frac{\alpha^2 (\sigma_{i}^2 + \sigma_{b}^2 - 2 \sigma_{i b})}{2\sigma_{i}^2 + 2 \alpha (\sigma_{i b} - \sigma_{i}^2) + C_2},
    \label{eq:psi_sigma}
\end{align}
where $i$, $o$, and $b$ denote the subimages of $I$, $O$, and $B$, respectively. For brevity, we omit the superscript $k$. We use the inverse SSIM because it is more convenient for iterative root finding.

By calculating the first and second derivatives of $\psi^k_\mu(\alpha)$ and $\psi^k_\sigma(\alpha)$, we can show that both derivatives are positive wherever their denominators are positive. For the valid blending range $\alpha \in [0,1]$, the denominator of $\psi^k_\mu$ is always positive. The denominator of $\psi^k_\sigma$ remains positive over this range when $\sigma_{i b} > -C_2/2$. We first consider $\sigma_{i b} > -C_2/2$, which is typically satisfied in practice. Under this condition, both components are positive, monotonically increasing, and convex (PMIC) over $[0,1]$, and inverse SSIM, $\psi^k(\alpha)$, is likewise PMIC. Because $\psi^k(\alpha)$ is well behaved, iterative root-finding methods perform robustly. When $\psi^k(1) \ge T_\mathrm{SSIM}^{-1}$, the optimal $\alpha$ is $1$. Otherwise, we can find the optimal $\alpha$ in a few iterations.
Let $\alpha_0=1$ and $\alpha_1<\alpha_0$ be two initial guesses. If $\psi^k(\alpha_1) > T_\mathrm{SSIM}^{-1}$, the secant method is expected to converge to the optimal $\alpha$ without oscillation. If $\psi^k(\alpha_1) < T_\mathrm{SSIM}^{-1}$, a false-position method is more appropriate. When $\sigma_{i b} \le -C_2/2$, we set $\alpha=0$ because the PMIC property does not hold and robust root finding becomes difficult.

We consider the noise-level dependency of $\psi^k_\mu(\alpha)$ and $\psi^k_\sigma(\alpha)$. Because the means are only weakly affected by noise, $\psi^k_\mu(\alpha)$ is expected to be relatively insensitive to noise level. For small subimages of noisy input, the input variance is dominated by noise and thus much larger than the output variance of the black-box denoiser. Therefore, we can expect that $\sigma_{i}^2 \gg \sigma_{b}^2$ and $\sigma_{i}^2 \gg |\sigma_{i b}|$, and we have $\psi^k_\sigma(\alpha) \simeq 1 + \alpha^2 / ( 2 - 2 \alpha )$. Note that this discussion is valid when $\alpha$ is away from $1$, which is our main interest. Thus, $\psi^k_\sigma(\alpha)$ is also approximately insensitive to noise level. For the small-noise case, the noise dependency is not as problematic as in the large-noise case because the black-box denoiser's output is already close to the input, and the blending process has less effect on the output. Therefore, we can expect that SSIM anchoring is more robust to noise-level variations than PSNR anchoring.

\section{Noise Distribution}
\label{sec:noise_distribution}

High PSNR or SSIM between the input and output images does not guarantee that the output appears natural to human observers. In this section, we discuss naturalness from the perspective of noise distribution. Let the ground-truth (GT) image be $G$. The residual noise is defined as $n_\mathrm{res} \equiv O - G$. When the distribution of $n_\mathrm{res}$ significantly deviates from a typical Gaussian shape, the output image $O$ is often perceived as unnatural. We use excess kurtosis $\gamma_2$ as a simple proxy for the deviation of residual noise from the assumed Gaussian noise distribution.

Because it depends on the GT image $G$, which is unknown, excess kurtosis is not a practical fidelity metric. However, it is useful as a naturalness measure in controlled experiments where $G$ is known.

\section{Experiments}

Given an input image $I$ and the output of a black-box denoiser $B$, we blend them to generate $O=I + \alpha \odot (B - I)$. Following the discussion in Section~\ref{sec:fidelity_metrics}, we can find the maximum feasible $\alpha$ that satisfies the fidelity condition, $\mathrm{PSNR}(I,O) \geq T_\mathrm{PSNR}$ or $\mathrm{SSIM}(I,O) \geq T_\mathrm{SSIM}$, for each subimage. Within this framework, we can balance fidelity and denoising performance for a wide range of black-box denoisers, including deep learning-based methods.

In this section, we demonstrate the effectiveness of the proposed framework through experiments. As input images, we use the DIV2K validation set \cite{Agustsson_2017_CVPR_Workshops,Timofte_2017_CVPR_Workshops}, which consists of 100 high-quality images. We add zero-mean Gaussian noise with standard deviations of 25, 50, and 75 to create noisy inputs. We then apply Real-ESRGAN \cite{wang2021realesrgan} as a black-box denoiser to obtain denoised outputs. We use Real-ESRGAN as a representative black-box restoration model because it performs both denoising and detail restoration on noisy inputs. For anchoring, we impose PSNR and SSIM constraints separately on each RGB channel: $\mathrm{PSNR}(I_{[RGB]},O_{[RGB]}) \ge T_\mathrm{PSNR}$ and $\mathrm{SSIM}(I_{[RGB]},O_{[RGB]}) \ge T_\mathrm{SSIM}$. We evaluate $T_\mathrm{PSNR} \in \{ 15, 17.5, 20, 22.5, 25, 27.5, 30 \}$ and $T_\mathrm{SSIM} \in \{0.3,0.4,0.5,0.6,0.7,0.8,0.9\}$. We use $7\times7$ windows for anchoring calculations. We also apply the OpenCV bilateral filter \cite{tomasi1998bilateral} with a conservative setting as a denoiser that produces natural-looking results but offers limited denoising performance. The filter parameters are $\mathrm{d}=11,\,\mathrm{sigmaColor}=25,\,\mathrm{sigmaSpace}=3$. While we have a closed-form solution for PSNR anchoring, SSIM anchoring requires iterative root finding, specifically the secant and false-position methods. We use a fixed number of iterations (4) for all experiments. Although 4 iterations are not necessarily optimal, they are sufficient to achieve the target SSIM values in our experiments. In the iterative process, we use $\alpha_0=1$ and $\alpha_1=0.75$ as initial guesses.

\begin{figure}[!htb]
    \centering
    \includegraphics[width=0.98\linewidth]{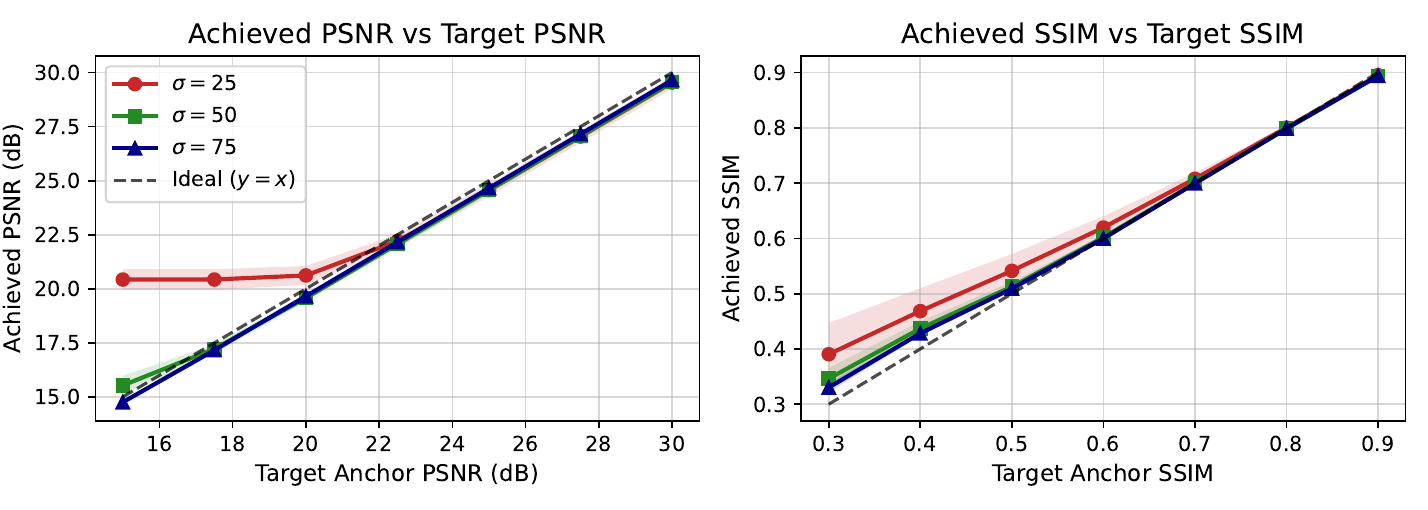}
    \caption{Achieved PSNR (left) and SSIM (right) values for different target values. (Real-ESRGAN)}
    \label{fig:anchoring_result}
\end{figure}

\begin{figure}[!htb]
    \centering
    \includegraphics[width=0.98\linewidth]{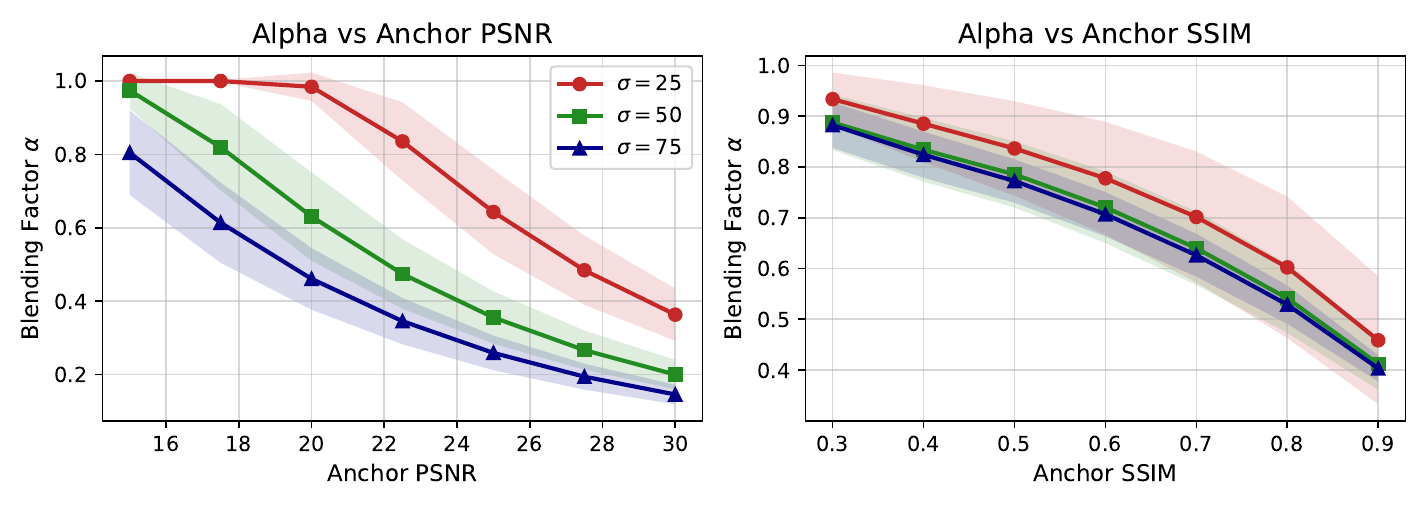}
    \caption{Blending factor $\alpha$ vs. anchor PSNR (left) and anchor SSIM (right). (Real-ESRGAN)}
    \label{fig:alpha_vs_anchor}
\end{figure}

\figurename~\ref{fig:anchoring_result} reports the means and standard deviations of the achieved PSNR and SSIM values for different targets. We also use $7\times7$ windows to compute local SSIM values for each RGB channel and then average them to obtain the global SSIM. These results confirm that the proposed framework effectively controls denoising fidelity, showing that the locally constant assumption for $\alpha$ is justified to an extent and a small number of iterations are sufficient. For lower target values, the achieved values deviate from the targets because the black-box denoiser outputs already satisfy the constraints. We also show the means and standard deviations of the resulting blending factors $\alpha$ in \figurename~\ref{fig:alpha_vs_anchor}. The blending factor $\alpha$ decreases as the target value increases, indicating that the output is more strongly anchored to the input image. As discussed in Section~\ref{sec:fidelity_metrics}, the blending factor $\alpha$ for PSNR anchoring is sensitive to noise level, while that for SSIM anchoring is relatively stable. This indicates that tuning the anchoring condition is easier for SSIM anchoring than for PSNR.

We evaluate denoising performance using PSNR between the GT image $G$ and the output image $O$. We use PSNR rather than SSIM because SSIM between GT and denoised outputs is often very low (e.g., $0.1$ or $0.2$), making cross-method comparisons less informative. In contrast, PSNR provides a straightforward pixel-wise fidelity measure. While it is a convenient metric, PSNR alone does not guarantee perceptual naturalness. Many deep learning-based denoisers achieve high PSNR but often produce unnatural artifacts. To investigate the perceptual naturalness of noise, we analyze the excess kurtosis of residual noise, $n_\mathrm{res} = O - G$, as discussed in Section~\ref{sec:noise_distribution}. For evaluation, PSNR is computed across the RGB channels, whereas the excess kurtosis $\gamma_2$ is evaluated on the luma (grayscale) residual noise, because human perception of noise is primarily governed by luma artifacts.

\begin{figure}[!htb]
    \centering
    \includegraphics[width=0.98\linewidth]{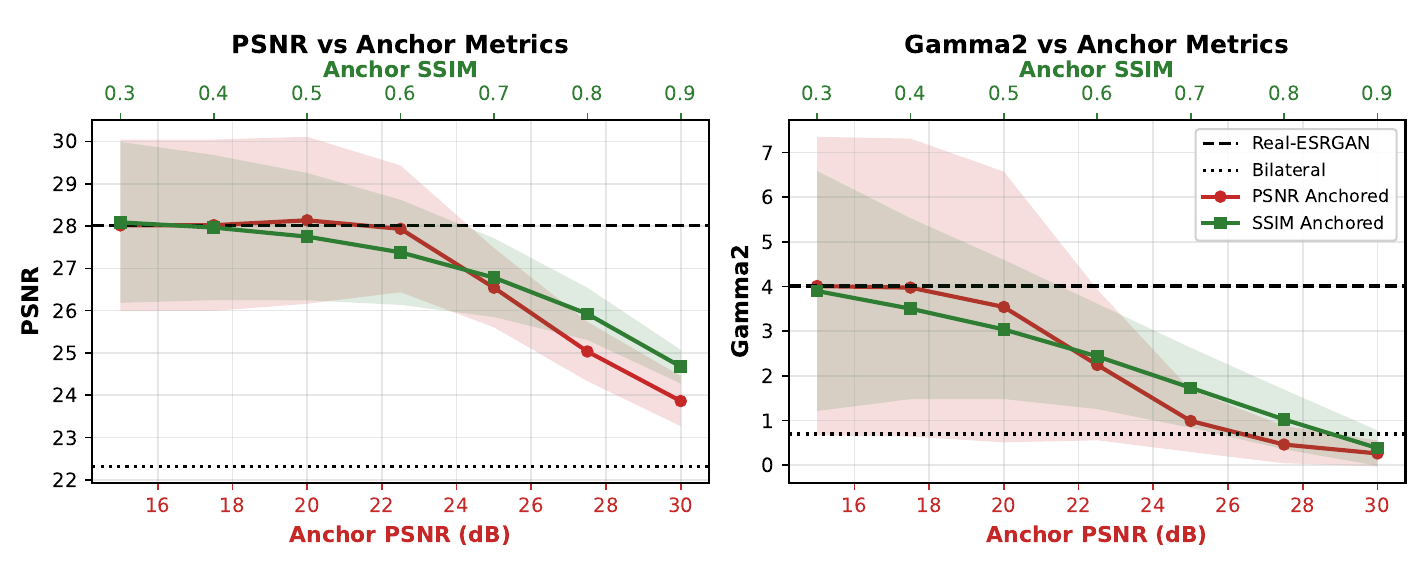}
    \includegraphics[width=0.98\linewidth]{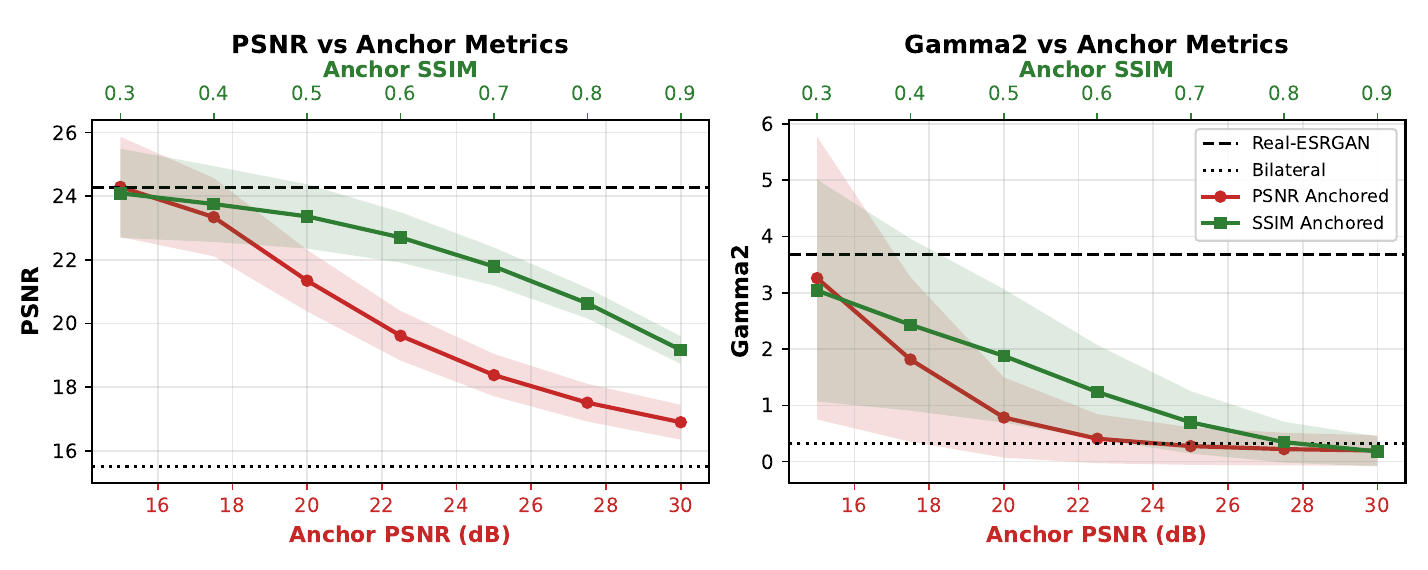}
    \includegraphics[width=0.98\linewidth]{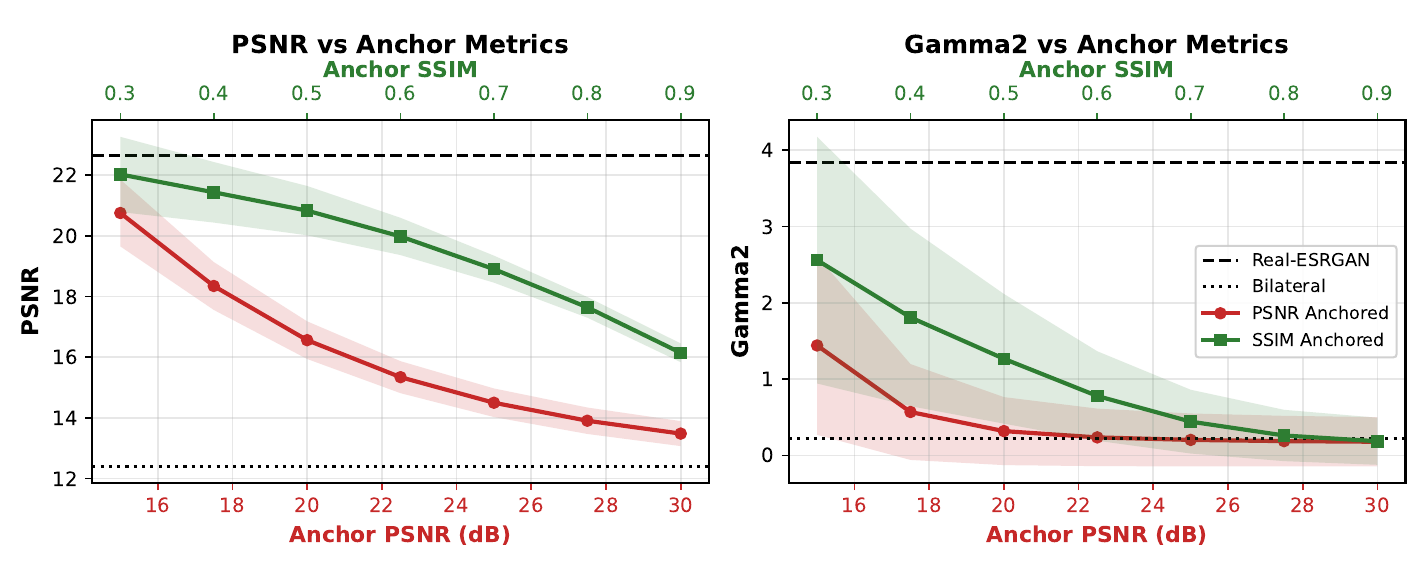}
    \caption{PSNR and excess kurtosis of the residual noise for different target values under Gaussian noise with $\sigma=25$ (top), $\sigma=50$ (middle), and $\sigma=75$ (bottom). Real-ESRGAN is used as a black-box denoiser.}
    \label{fig:graph}
\end{figure}

\figurename~\ref{fig:graph} (top: $\sigma=25$, middle: $\sigma=50$, bottom: $\sigma=75$) shows PSNR and excess kurtosis for different target values\footnote{For visual results, please refer to the supplementary material. Our framework finds the balance between the denoising strength and fidelity without severe artifacts.}. These results indicate that fidelity control provides a practical way to balance denoising performance and statistical naturalness. For $\sigma=25$, PSNR anchoring even achieves higher PSNR than Real-ESRGAN; we leave detailed analysis of this behavior for future work. The two anchoring strategies exhibit clearly different trends. Across $\sigma=25$, $\sigma=50$, and $\sigma=75$, SSIM anchoring shows similar PSNR-kurtosis trade-off curves, and $T_\mathrm{SSIM}=0.8$ is a reasonable operating point for all three noise levels, yielding sufficiently high PSNR (compared to bilateral) and low excess kurtosis. In contrast, PSNR anchoring behaves differently across noise levels. Although $T_\mathrm{PSNR}=25$ yields low excess kurtosis in all cases, the achieved PSNR is relatively high for $\sigma=25$ but low for $\sigma=50$ and $\sigma=75$. Thus, PSNR anchoring is sensitive to noise level, while SSIM anchoring is relatively stable. This noise dependency of PSNR anchoring is consistent with the previous analysis.

\begin{figure}[!htb]
    \centering
    \includegraphics[width=0.98\linewidth]{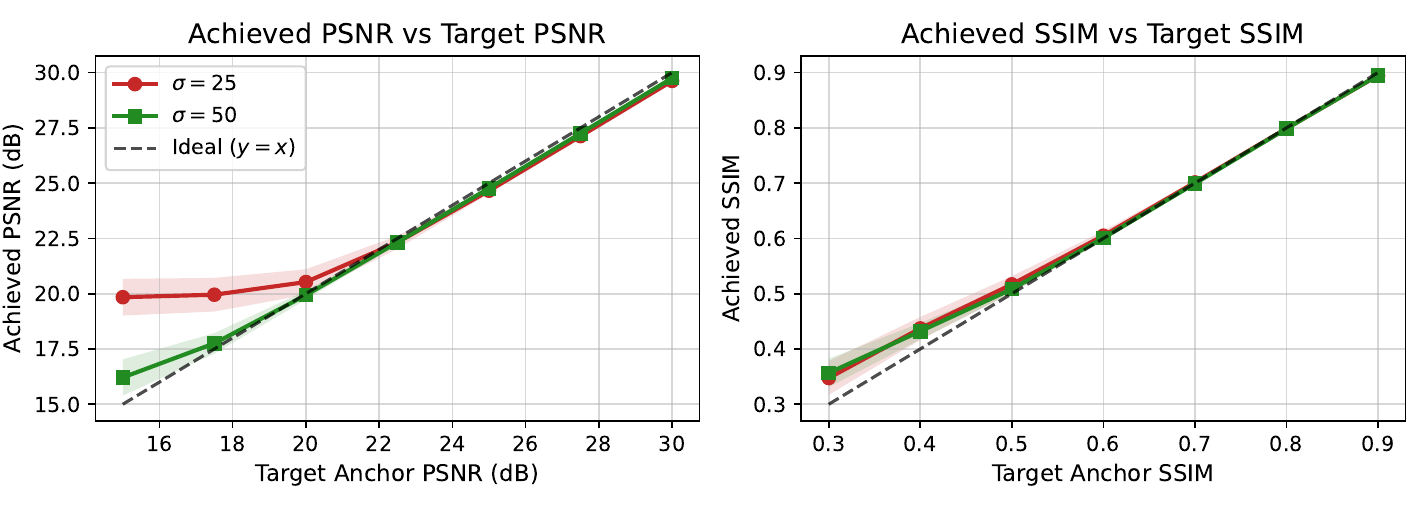}
    \caption{Achieved PSNR (left) and SSIM (right) values for different target values. (Non-local Means)}
    \label{fig:anchoring_result_fnm}
\end{figure}

\begin{figure}[!htb]
    \centering
    \includegraphics[width=0.98\linewidth]{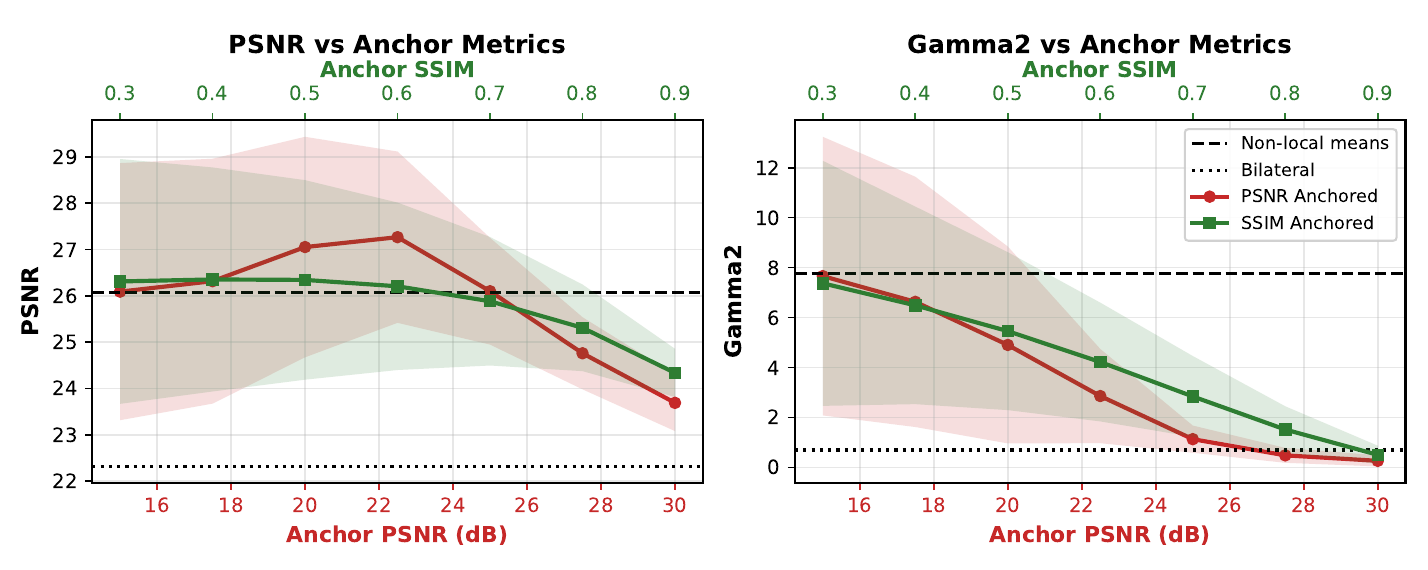}
    \includegraphics[width=0.98\linewidth]{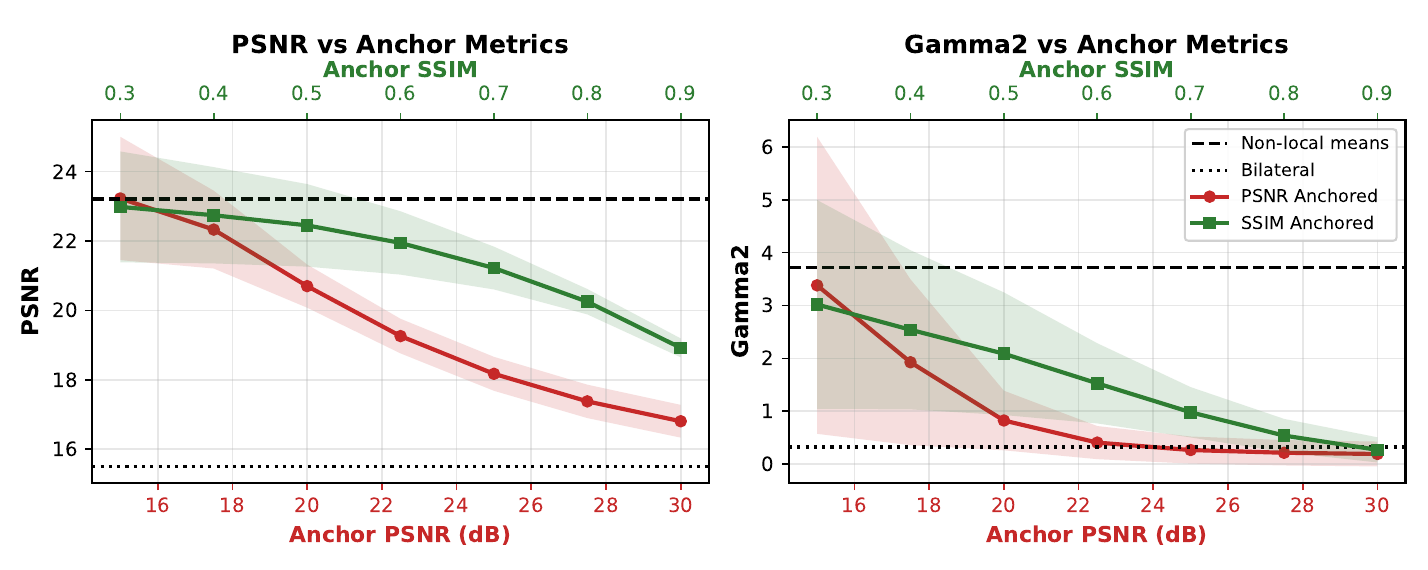}
    \caption{PSNR and excess kurtosis of the residual noise for different target values under Gaussian noise with $\sigma=25$ (top) and $\sigma=50$ (bottom). Non-local means is used as a black-box denoiser.}
    \label{fig:graph_fnm}
\end{figure}

To further validate these findings, we conduct additional experiments using the non-local means denoiser \cite{buades2005non} as another black-box denoiser. We use the OpenCV implementation, \texttt{fastNlMeansDenoisingColored}, with parameters $\mathrm{h}=25$, $\mathrm{hColor}=25$, $\mathrm{templateWindowSize}=7$, and $\mathrm{searchWindowSize}=21$. \figurename~\ref{fig:anchoring_result_fnm} shows the anchoring results for different conditions, indicating that PSNR and SSIM anchoring also work well for non-local means denoising. \figurename~\ref{fig:graph_fnm} (top: $\sigma=25$, bottom: $\sigma=50$) shows the corresponding PSNR and excess kurtosis of the residual noise. We report only the $\sigma=25$ and $\sigma=50$ cases to save space. The same trend is observed: SSIM anchoring exhibits more consistent behavior across noise levels than PSNR anchoring, suggesting that this behavior is not specific to Real-ESRGAN. Thus, tuning the anchoring condition is easier for SSIM anchoring than for PSNR.

\section{Conclusion}

In this paper, we proposed a framework that anchors the output of a black-box denoiser to the input image using fidelity metrics. We demonstrated that local blending between denoiser output and input image enables explicit control of fidelity metrics such as PSNR and SSIM. Experiments showed that this approach effectively balances denoising performance and statistical naturalness, measured by excess kurtosis of residual noise. In particular, SSIM anchoring exhibits more consistent behavior across noise levels, making it a robust choice for practical applications. The framework treats the denoiser as a black-box mapping and does not require modification or retraining.

Future work includes exploring additional fidelity metrics and metric combinations, and extending the framework to more complex noise models and real-world scenarios. Importantly, the proposed strategy of blending black-box outputs with inputs under explicit fidelity constraints is generic and can be extended to other image restoration tasks, such as super-resolution and deblurring.

A limitation of our framework is that, because it is based on linear blending, performance degrades when the denoiser output differs substantially from the input or when the input noise is so strong that the original image structure is heavily corrupted.

\section*{Acknowledgment}

The author would like to thank Takeshi Miura for insightful discussions, and Masaki Hilaga for his support and authorization to conduct this research as part of corporate duties. An AI language model was used to assist with the manuscript's structure and English phrasing.

\bibliographystyle{IEEEtran}
\bibliography{references}

\end{document}